\documentclass[sigconf]{acmart}

\AtBeginDocument{%
  }

\setcopyright{acmlicensed}
\copyrightyear{2024}
\acmYear{2024}
\acmDOI{XXXXXXX.XXXXXXX}
\usepackage{algpseudocode}
\usepackage{algorithm}
\usepackage{appendix}
\usepackage{float}
\usepackage{graphicx}

\acmConference{Preprint Paper}
\acmISBN{978-1-4503-XXXX-X/18/06}

\begin{document}

\title{Improving Mixed-Criticality Scheduling with Reinforcement Learning}

\author{Muhammad El-Mahdy}
\affiliation{%
  \institution{The American University in Cairo}
  \city{Cairo}
  \country{Egypt}}
\email{muhammadahmedelmahdy@aucegypt.edu}
\author{Nourhan Sakr}
\affiliation{%
  \institution{The American University in Cairo}
  \city{Cairo}
  \country{Egypt}}
\email{n.sakr@columbia.edu}

\author{Rodrigo Carrasco}
\affiliation{%
  \institution{Pontificia Universidad Católica de Chile}
  \city{Santiago}
  \country{Chile}}
  \email{rcarrass@uc.cl}



\begin{abstract}
This paper introduces a novel reinforcement learning (RL) approach to scheduling mixed-criticality (MC) systems on processors with varying speeds. Building upon the foundation laid by \cite{6728862}, we extend their work to address the non-preemptive scheduling problem, which is known to be NP-hard. By modeling this scheduling challenge as a Markov Decision Process (MDP), we develop an RL agent capable of generating near-optimal schedules for real-time MC systems. Our RL-based scheduler prioritizes high-critical tasks while maintaining overall system performance.

Through extensive experiments, we demonstrate the scalability and effectiveness of our approach. The RL scheduler significantly improves task completion rates, achieving around 80\% overall and 85\% for high-criticality tasks across 100,000 instances of synthetic data and real data under varying system conditions. Moreover, under stable conditions without degradation, the scheduler achieves 94\% overall task completion and 93\% for high-criticality tasks. These results highlight the potential of RL-based schedulers in real-time and safety-critical applications, offering substantial improvements in handling complex and dynamic scheduling scenarios.
\end{abstract}


\received{07 June 2024}

\maketitle

\section{Introduction}
In modern computing systems, real-time scheduling plays a pivotal role in ensuring the execution and completion of tasks across different applications and industries. Central to many real-time mixed-criticality (MC) systems, where tasks vary in terms of importance and criticality within the same computing environment. In such systems, tasks are usually associated with defined criticality levels, representing the severity of consequences resulting from the failure to execute respective tasks. For example, in aircraft control systems, tasks such as sensor data processing, flight control commands, and collision avoidance are highly critical, as missing their deadlines can lead to catastrophic outcomes. Conversely, in multimedia processing, tasks like video rendering and audio synchronization are less critical; missing their deadlines results in performance degradation but poses no immediate safety risk. This illustrates the varying criticality of tasks within the same computing environment.

The challenge of mixed-criticality scheduling lies in efficiently allocating computing resources to tasks with different criticality levels, prioritizing critical tasks, and ensuring their completion. Achieving this delicate balance requires the development of sophisticated scheduling algorithms capable of dynamically adapting to changing system conditions and accommodating the diverse requirements of mixed-criticality tasks. Mixed criticality (MC) scheduling presents a complex and dynamic challenge, primarily because MC systems typically operate across multiple modes during runtime, often corresponding to different criticality levels. For example, in a dual-criticality system running on a variable-speed processor, the system can transition between a "normal" mode and a "degradation" mode in response to external factors such as heat, high supply voltage, or fluctuations in clock frequency, leading to a decrease in processor speed, i.e., degradation of processor performance.

\paragraph{Our Contribution:}
In this research, we propose an offline scheduler using Reinforcement learning to schedule dual-criticality jobs on a varying speed processor. We model the non-preemptive problem, known to be NP-hard, using a Markov Decision Process (MDP) and train different RL agents to schedule the tasks correctly and prioritize HI-critical tasks. Our approach demonstrates good results when it comes to tackling this NP-hard problem in an effective and scalable way. Finally, we use both synthetic and real data to test our results. 

The next sections of this paper start with the related work in \S 2, then the detailed methodology in \S 3, the data approach in \S 4, the experimental results in \S 5 along with sensitivity analysis in \S 6, and finally conclusing and presenting future work in \S 7. We believe that RL is a good approach for this problem due to its high scalability, ability to solve complex dynamic problems, and ability to solve an NP-Hard problem like offline non-preemptive MC scheduling.

\section{Related Work}
\subsection{Approaches to MC Scheduling}
Since the publishing of the white paper by Anderson et al., researchers have been able to understand more about MC systems and the significant impact of criticality levels on scheduling approaches aiming to manage them. Accordingly, researchers start solving this problem using combinatorial or  mathematical  optimization approaches. For instance, the authors in \cite{10.1145/3566097.3567851} show the scheduling strategies for real-time systems with fault tolerance requirements. Their contribution tackels an MC system with integer multiple WCETs. Such a model introduces some constraints on task execution times to be multiples of the lowest-criticality WCET. Their model simplifies scheduling analysis and improves the speedup bound for the EDF-VD algorithm. The authors also suggest dropping the relations model, which drops tasks based on overruns, to give fine-grained control over task behavior. While such models give promising advancements in real-time scheduling, current research shows the need for further investigation into better scheduling algorithms. More recently, Baruah and Guo \cite{6728862} model the problem of offline preemptive MC scheduling using linear programming (LP) and solve for the schedule that is able to tolerate the minimum degradation speed.

\subsection{Reinforcement learning in Scheduling problems}
Reinforcement learning (RL) has greatly emerged as an AI field where agents are trained to make sequential decisions and maximize the reward based on the actions taken. This AI field has been used various times in literature to solve dynamic scheduling problems. In \cite{LIU2023106294}, the work presents a deep multi-agent reinforcement learning (MARL) approach for addressing the complexities of dynamic job shop scheduling problems. By going through a 3-stage experimentation process, they devised a methodology that they evaluated against existing benchmarks. They then show the significant performance enhancements that they introduced to components like the DQN, reward shaping, and parameter sharing. Their study also does an evaluation on static problem instances and could show competitive performance leading to a genetic algorithm on larger problem instances. On the other hand, the authors in \cite{shyalika2020reinforcement} provide a comprehensive overview of the RL techniques for dynamic scheduling algorithms. They start by tackling model-free RL, including Q learning and DQN, and how they have a significant application in real-time decision-making scenarios, including job scheduling. On the other hand, model-based RL leverages the learned model from the environment to plan actions and simulate the future states accordingly. They also show both integrations in the Dyna-Q algorithm and present an intriguing room for highlighting the uncertain nature of real-world practical environments. They finally highlight some frameworks that use RL for scheduling like Ordinal Sharing Learning (OSL), Gossip-Based Reinforcement Learning (GRL) and Centralized Learning Distributed Scheduling (CLDS). Also, in \cite{NEURIPS2021_1e4d3617}, the authors discuss the powerful model-based RL techniques and how they can enhance efficiency. The paper addresses how the parameters and data play an important role in policy training and shows how we can increase the data to improve performance. They propose a framework to schedule the hyperparameters, including the real data ratio while training model-based policy optimization algorithms. Overall, this paper underscores the criticality of effective hyperparameter scheduling in model-based RL and offers a practical solution in the form of AutoMBPO, paving the way for improved performance and efficiency in real-world RL applications. Additionally, the paper \cite{grinsztajn2021readys}, proposes an innovative approach of READYS, a reinforcement learning algorithm that combines both graph convolutional networks and actor-critic algorithms for dynamic scheduling. They model the computations as directed acyclic graphs and demonstrate the superior performance of their approach in handling the uncertainty and generalization across the different tasks.

\subsection{Our scope}
Since most of the work done on the MC scheduling problem was done using combinatorial approaches, the problem of the NP Hard non preemptive version arises. In this research, we believe that reinforcement learning, being used in other dynamic scheduling problems, can be a suitable solution to this problem given its ability to solve the NP hard problems, providing a scalable solution that deals with dynamic complex problems. To the best of our knowledge, little work has been done using Reinforcement learning on the MC scheduling. The authors in \cite{sakr2021scheduling} have managed to get preliminary results to this problem using reinforcement learning. In this work, we build on the potential of deep reinforcement learning in other scheduling problems and the prior work done in this problem and provide an RL scheduling agent for MC scheduling. We do extensive experiments trying to get a more scalable solution that solves the offline non preemptive problem. Without loss of generality, we assume a dual criticality system that only has two criticality levels). The agent is able to schedule the tasks offline based on information of predetermined future tasks, along with their release times, deadlines, worst case execution times (WCETs) and criticality levels, thereby providing a scalable and better solution on data simulated to mimic real data.

\section{Methodology}
We have a job instance such that it consists of n independent job, and the RL agent's task is to schedule the jobs in this instance in the correct way, prioritizing the near laxity HI priority jobs, the near laxity LO priority jobs, the far laxity HI priority and finally the far laxity LO priority jobs, in order. The decision of whether the job has near laxity or far laxity is based on a dynamic threshold. Please refer to section \ref{sec:Appendix} to see how this threshold is determined. The reason behind this is to find the balance between the HI job completion rate and the overall completion rate. We define a job to model tasks within the system, such that each job is characterized by its release time, deadline, processing time, and criticality level. Upon initialization, the job's remaining processing time is set to its initial processing time, and its laxity is calculated as the difference between its deadline and processing time. We use various flags to monitor the status of the job, like released, executed, starved, and scheduled, to track the job's state throughout its life cycle. Additionally, the job's status is updated based on the current time frame, taking various parameters like release time, execution, and deadline into account.

\subsection{Offline MC scheduling}
We model the problem of offline non-preemptive scheduling as a Markov decision process(MDP), in which the agent takes action to maximize his reward function. Since it's an offline problem, in this MDP, the agent receives an observation with \textbf{all the jobs} from the beginning, including all the job attributes that were mentioned earlier. Each observation step is updated, where the agent keeps receiving all the jobs but with their updated attributes based on the time taken now. Each time step, the agent takes an action from a discrete action space with actions equal to the number of jobs, and the action is the job the agent chooses to schedule at this time step. The agent is then rewarded or penalized based on the action he takes. Please see section \ref{sec:Appendix} for the detailed design and implementation of the MDP and the reward function. The scheduling environment, i.e., the MDP, is initialized with time 0 and a speed 1, assuming that the processor is in normal mode. We then model the degradation by having a parameter in the environment for the \textbf{the degradation chance}. Every time step, we employ an algorithm that determines whether degradation occurs based on a probability threshold. If degradation occurs, it assigns the object's speed to a random value between the lowest possible degraded speed and full speed. If no degradation occurs, the object's speed remains full (1). We employ \textbf{masked PPO} RL algorithm leveraging the idea of action masking by masking the jobs that can't be scheduled, i.e., not released, starved, or already executed. Additionally, given that we are giving priority to the HI jobs during degradation, we mask the LO jobs if and only if the processor degrades and there are available HI jobs, so we force the agent to choose a HI criticality job, and the agent chooses based on the deadline and the near laxity. For the MDP used for offline non-preemptive scheduling, the MDP is modeled as follows:-
\begin{itemize}
    \item  States: Observed job characteristics, such as release time, deadline, processing time, criticality level, remaining processing time, laxity, and laxity rank
    \item Actions: Selection of a job to be scheduled
    \item Rewards: Based on job criticality level and laxity rank
    \item Transition probabilities: Deterministic (actions directly affect the environment state)
\end{itemize}

\subsection{Degradation Modeling}
In this work, we simulated the stochastic degradation of the processor using a probabilistic approach to modeling the degradation. Initially, a random number between 0 and 1 is generated, and then if this number is below the degradation threshold, then the object speed is reduced to a random value between the lowest speed that keeps the instance feasible and 1 if the random number is above the degradation threshold, then the processor remains at the full speed (1). We determine the degradation threshold in every training episode by drawing a random number from a uniform distribution between 0.05 and 0.95. We assume that the RL agent is accompanied by a degradation forecasting model, and for simplicity, we use the uniform distribution model to get the processor speed at the different time steps.

\section{Data modeling}
As mentioned earlier, the job instance is modeled as a set of n independent jobs. We define a job to model tasks within the system, such that each job is characterized by its release time, deadline, processing time, and criticality level. Upon initialization, the job's remaining processing time is set to its initial processing time, and its laxity is calculated as the difference between its deadline and processing time. We use various flags to monitor the status of the job, like released, executed, starved, and scheduled, to track the job's state throughout its life cycle. Additionally, the job's status is updated based on the current time frame, taking various parameters like release time, execution, and deadline into account. We generate synthetic job instances on the fly to train the RL agent. The data generated for training is similar, in distribution, to real data coming from a server in the dataset studied in \cite{Carrasco2018:rcasEjor}, which will be used later for testing the RL agent, together with testing on synthetic data. 

\subsection{Summary of the data from the server}
Based on the statistical analysis of attributes like release time and processing time, the server data can be categorized into 5 groups.
Tables 1, 2, 3, 4, and 5 show the statistics of the five groups. To better represent the data, we scale the attributes by dividing the release and processing times by 1,000. We also decide on a threshold to determine if the job has criticality level 0, i.e., LO, or 1, i.e., HI. Since the dataset was missing job deadlines, we simulated deadlines so that the job remained feasible in the same way we generated the synthetic data.

\begin{table}[!htbp]
\renewcommand{\arraystretch}{1.0}
\caption{Statistics of group 1 jobs of Release Time, Deadline, Processing Time, and Criticality Level}
\label{table_statistics_1}
\centering

\begin{tabular}{p{0.09\textwidth}|p{0.05\textwidth}|p{0.07\textwidth}|p{0.07\textwidth}|p{0.05\textwidth}}
\hline
 & Release Time  & Processing Time & Criticality Level \\
\hline\hline
Count & 994 & 994 & 994 \\
Mean & 6643  & 118522 & 7 \\
Standard Deviation & 13421  & 592148 & 4 \\
Minimum & 0  & 1 & 0 \\
Maximum & 84809  & 10058893 & 10 \\
\hline
\end{tabular}
\end{table}
\begin{table}[!htbp]
\renewcommand{\arraystretch}{1.3}
\caption{Statistics of group 2 jobs of Release Time, Deadline, Processing Time, and Criticality Level}
\label{table_statistics_2}
\centering
\begin{tabular}{p{0.09\textwidth}|p{0.05\textwidth}|p{0.07\textwidth}|p{0.07\textwidth}|p{0.05\textwidth}}
\hline
 & Release Time  & Processing Time & Criticality Level \\
\hline\hline
Count & 2436  & 2436 & 2436 \\
Mean & 8437  & 81050 & 7 \\
Standard Deviation & 14431  & 262915 & 4 \\
Minimum & 0  & 1 & 0 \\
Maximum & 85049  & 3655203 & 10 \\
\hline
\end{tabular}
\end{table}

\begin{table}[!htbp]
\renewcommand{\arraystretch}{1.3}
\caption{Statistics of group 3 jobs of Release Time, Deadline, Processing Time, and Criticality Level}
\label{table_statistics_3}
\centering
\begin{tabular}{p{0.09\textwidth}|p{0.05\textwidth}|p{0.07\textwidth}|p{0.07\textwidth}|p{0.05\textwidth}}
\hline
 & Release Time  & Processing Time & Criticality Level \\
\hline\hline
Count & 7209 & 7209 & 7209 \\
Mean & 11425  & 46411 & 6 \\
Standard Deviation & 17729  & 222446 & 5 \\
Minimum & 0 & 1 & 0 \\
Maximum & 86291  & 11379765 & 10 \\
\hline
\end{tabular}
\end{table}

\begin{table}[!htbp]
\renewcommand{\arraystretch}{1.3}
\caption{Statistics of group 4 jobs of Release Time, Deadline, Processing Time, and Criticality Level}
\label{table_statistics_4}
\centering
\begin{tabular}{p{0.09\textwidth}|p{0.05\textwidth}|p{0.07\textwidth}|p{0.07\textwidth}|p{0.05\textwidth}}
\hline
 & Release Time  & Processing Time & Criticality Level \\
\hline\hline
Count & 31055  & 31055 & 31055 \\
Mean & 14226  & 28371 & 4 \\
Standard Deviation & 18843  & 128906 & 5 \\
Minimum & 0 & 1 & 0 \\
Maximum & 85315  & 5920238 & 10 \\
\hline
\end{tabular}
\end{table}
\begin{table}[!htbp]
\renewcommand{\arraystretch}{1.3}
\caption{Statistics of group 5 jobs of Release Time, Deadline, Processing Time, and Criticality Level}
\label{table_statistics_5}
\centering
\begin{tabular}{p{0.09\textwidth}|p{0.05\textwidth}|p{0.07\textwidth}|p{0.07\textwidth}|p{0.05\textwidth}}
\hline
 & Release Time  & Processing Time & Criticality Level \\
\hline\hline
Count & 1266780  & 1266780 & 1266780 \\
Mean & 23422  & 2981 & 3 \\
Standard Deviation & 24420 & 32443 & 3 \\
Minimum & 0  & 1 & 0 \\
Maximum & 86388  & 3279612 & 10 \\
\hline
\end{tabular}
\end{table}

\subsection{Data generation approach}
Our data generation approach tries to simulate the real data described earlier such that these data occur in task-based scheduling environments. We start generating training data on the fly each training episode such that the data generation is monitored by a parameter \textbf{LO}, which is the percentage of low jobs in the instance. Since the real data distribution is exponential, we generate the jobs' release time by sampling a random value from an exponential distribution and similarly for the deadline. We then validate that each job's instance and data generated are valid. This validation is done by ensuring that the job remains within the bounds of the release time and the deadline. We then filter the instances based on the feasibility and use the Early Deadline First (EDF) policy to modify the deadline such that all the instances have at least 1 schedule that ensures the completion of all the jobs under no degradation using EDF. 

\subsection{Evaluation metrics}
Our evaluation of the system depends on three aspects, which are 
\begin{itemize}
    \item Average reward value: Used to monitor if the agent is learning while training
    \item Job completion and average number of missed jobs: Used during testing to see how many jobs an agent completes and how many it misses.
    \item Speed: used to evaluate the schedules using an algorithm discussed in section \ref{sec:Appendix} for deciding on the lowest degradation speed that keeps the HI jobs completion rate.
\end{itemize}

\section{Experimental results}
We started doing experiments using the MDPs described in the methodology section. We initially monitor \textbf{the average reward value} for each training episode. Then, when testing the trained agents, we use \textbf{Job completion rate and CPU speed} as evaluation metrics to monitor the system's performance. We train an agent on the MDP described earlier and the policy, \textbf{Masked PPO}, where the agent can schedule up to 50 jobs.

After designing the MDP that was mentioned earlier for the offline non-preemptive scheduling, we start training the RL agent using \textbf{Masked PPO policy}. We evaluate the agent based on the two items below:-
\begin{itemize}
    \item \textbf{Average reward value: } We monitored the average reward value, showing that it increased through the different time steps. The way they are increasing shows that the agent is learning to do the scheduling correctly, and the results show that the agent is behaving as expected. Figure 1 shows how the episode reward mean was increased during training the agent until it reached a certain level.
    \item \textbf{Job completion: } We evaluate the job completion based on the percentage of the HI jobs completed and the overall jobs completed. Since, as mentioned earlier, the data generation approach ensures that the instance has at least 1 schedule that completes all the jobs, our benchmark is the full completion of the jobs, and we monitor how many jobs are missed. 
    \item \textbf{Speed: } We evaluate the schedule proposed by the agent by seeing how much degradation the CPU can degrade to and whether the HI criticality jobs still get completed. Please see section \ref{sec:Appendix} for the speed evaluation algorithm.
\end{itemize}
\begin{figure}[t]
  \centering
  \includegraphics[width=1\linewidth]{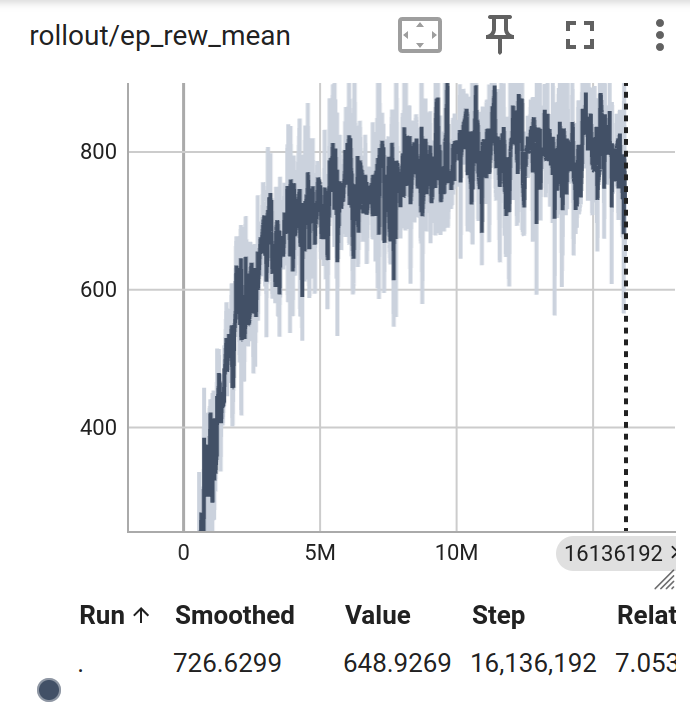}
  \caption{The training curve of the agent using masked PPO policy. The results show how the agent was learning with more time steps based on the rewards given to it that are described in the rewards function.}
  \label{fig:example}
\end{figure}

Our initial experiments test the RL scheduler on a CPU that doesn't degrade, i.e., speed=1, and then we start testing on a varying-speed processor.
\subsection{No degradation}
\subsubsection{Testing on synthetic data}
We start by testing the agent with 100000 instances, where every instance consists of 50 jobs, and the percentage of low jobs in every instance is decided by a random value drawn from a uniform distribution ranging from 0 to just below 1, specifically from 0 to \(1 - \frac{2}{\text{num\_jobs}}\), ensuring variability while avoiding values too close to 1.
Table 6 shows the results after doing this experiment:
\begin{table}[!htbp]
\renewcommand{\arraystretch}{1.3}
\caption{The results of the experiment of 100000 instances of synthetic data}
\label{table_statistics_5}
\centering
\begin{tabular}{l|r|r}
\hline
\multicolumn{1}{c|}{} & \multicolumn{2}{c}{Statistics} \\
\hline\hline
 & Average job completion rate & Average missed jobs\\
\hline
HI jobs & 92.83\% & 1.49 jobs \\
Overall jobs & 94.16\% & 2.91 jobs \\

\hline
\end{tabular}

\end{table}
\newline
\subsubsection{Testing on real server data}
As described earlier, we tested our RL agent using the first two groups of server data. Given the size of the instances in these two groups, we used the agent trained on scheduling 50 jobs. 

\paragraph{Group 1}
The first group consisted of 125 instances, and the number of jobs in this instance ranged from 4 to 13 jobs. Since the agent is initially trained on scheduling 50 jobs, we fill the remaining places with dummy jobs and mask these dummy jobs by 0 to prevent the agent from choosing these jobs.  To get an aggregate result for this group, we analyze the average completion rate and the average number of missed jobs for the HI criticality jobs and overall. Table 7 shows the results of this experiment. In addition, we evaluated the average speed of this group using the speed algorithm explained earlier. The results showed that, on average, the processor can degrade to \textbf{50.66 \%} and still maintain the HI criticality job completion rate.

\begin{table}[!htbp]
\renewcommand{\arraystretch}{1.3}
\caption{The results of the experiment of group1 of the server data}
\label{table_statistics_5}
\centering
\begin{tabular}{l|r|r}
\hline
\multicolumn{1}{c|}{} & \multicolumn{2}{c}{Statistics} \\
\hline\hline
 & Average job completion rate & Average missed jobs\\
\hline
HI jobs & 65\% & 1.52 jobs \\
Overall jobs & 70 \% & 2.456 jobs \\

\hline
\end{tabular}

\end{table}
\paragraph{Group 2}
The second group consisted of 109 instances where the number of jobs in this instance ranged from 14 to 35 jobs. Since the agent is initially trained on scheduling 50 jobs, we fill the remaining places with dummy jobs and mask these dummy jobs by 0 to prevent the agent from choosing these jobs. Table 8 shows the results of this experiment.  Upon evaluating the average speed of the schedule, we found that the processor can degrade to \textbf{52.30\% }and still maintain its HI criticality job completion rate. 
\begin{table}[!htbp]
\renewcommand{\arraystretch}{1.3}
\caption{The results of the experiment of group2 of the server data}
\label{table_statistics_5}
\centering
\begin{tabular}{l|r|r}
\hline
\multicolumn{1}{c|}{} & \multicolumn{2}{c}{Statistics} \\
\hline\hline
 & Average job completion rate & Average missed jobs\\
\hline
HI jobs & 68\% & 2.7 jobs \\
Overall jobs & 66 \% & 7.3 jobs \\
\hline
\end{tabular}

\end{table}

\subsection{Varying speed processor}
In this section, we repeat the previous experiments but on a processor of varying speed, assuming that degradation occurs at different times.
\subsubsection{Testing on synthetic data}
Doing the experiment on 100,000 instances with varying percentages of LO jobs and varying degradation thresholds, the results of our system showed decent performance, though it was less than the performance with no degradation.  However, this makes sense, as degradation prevents the agent from completing all the jobs required. Each instance has 50 jobs, and Table 9 shows the results of these experiments.

\begin{table}[!htbp]
\renewcommand{\arraystretch}{1.3}
\caption{The results of the experiment of synthetic data on varying speed processor}
\label{table_statistics_5}
\centering
\begin{tabular}{l|r|r}
\hline
\multicolumn{1}{c|}{} & \multicolumn{2}{c}{Statistics} \\
\hline\hline
 & Average job completion rate & Average missed jobs\\
\hline
HI jobs & 85.34\% & 3.2 jobs \\
Overall jobs & 80.65 \% & 9.6 jobs \\

\hline
\end{tabular}

\end{table}

\subsubsection{Testing on real server data}
We repeat the previous experiments but on a processor of varying speed. Figures 2 and 3 show example schedules from Groups 1 and 2, respectively.

\begin{figure}[htbp]
  \centering
  \includegraphics[width=1\linewidth]{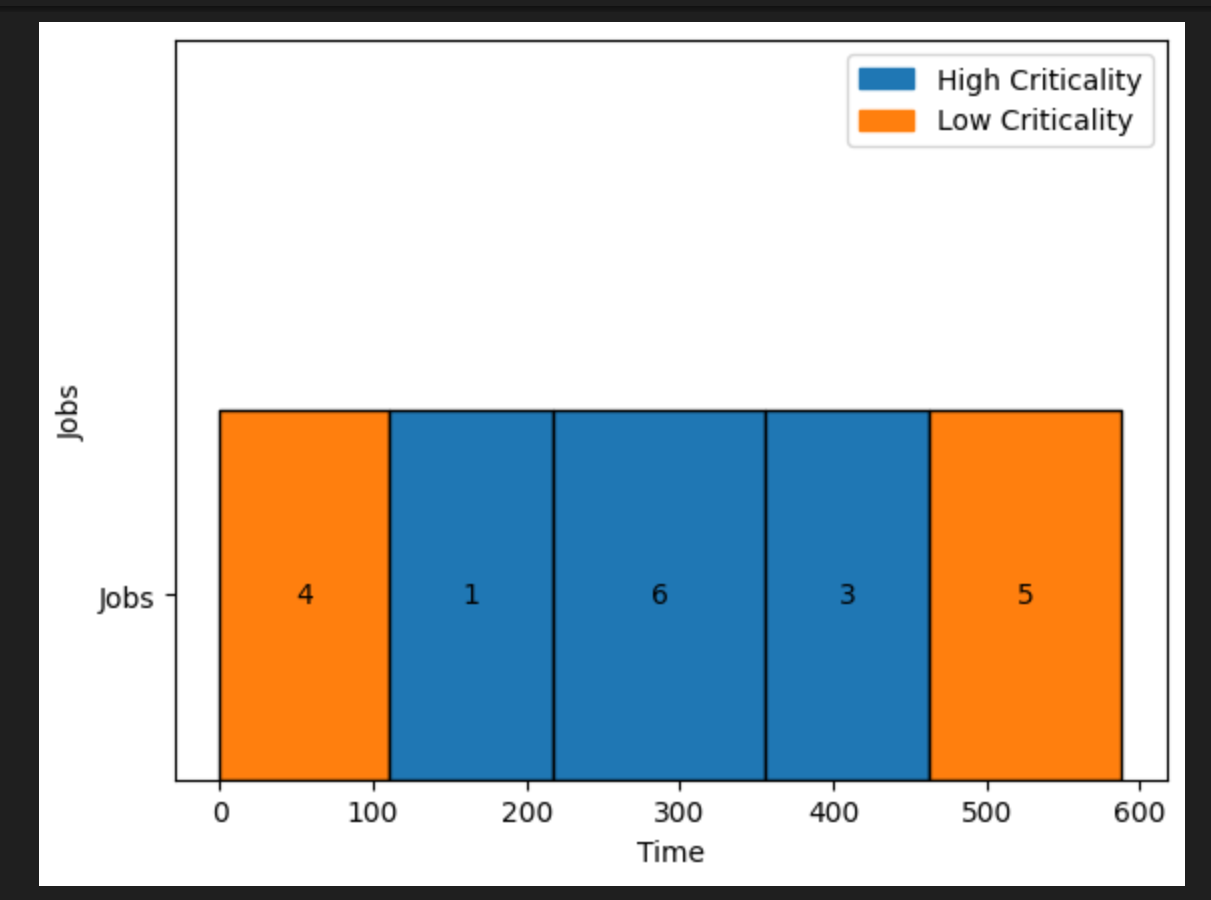}
  \caption{the schedule of an instance from Group 1}
  \label{fig:example}
\end{figure}

\begin{figure}[htbp]
  \centering
  \includegraphics[width=1\linewidth]{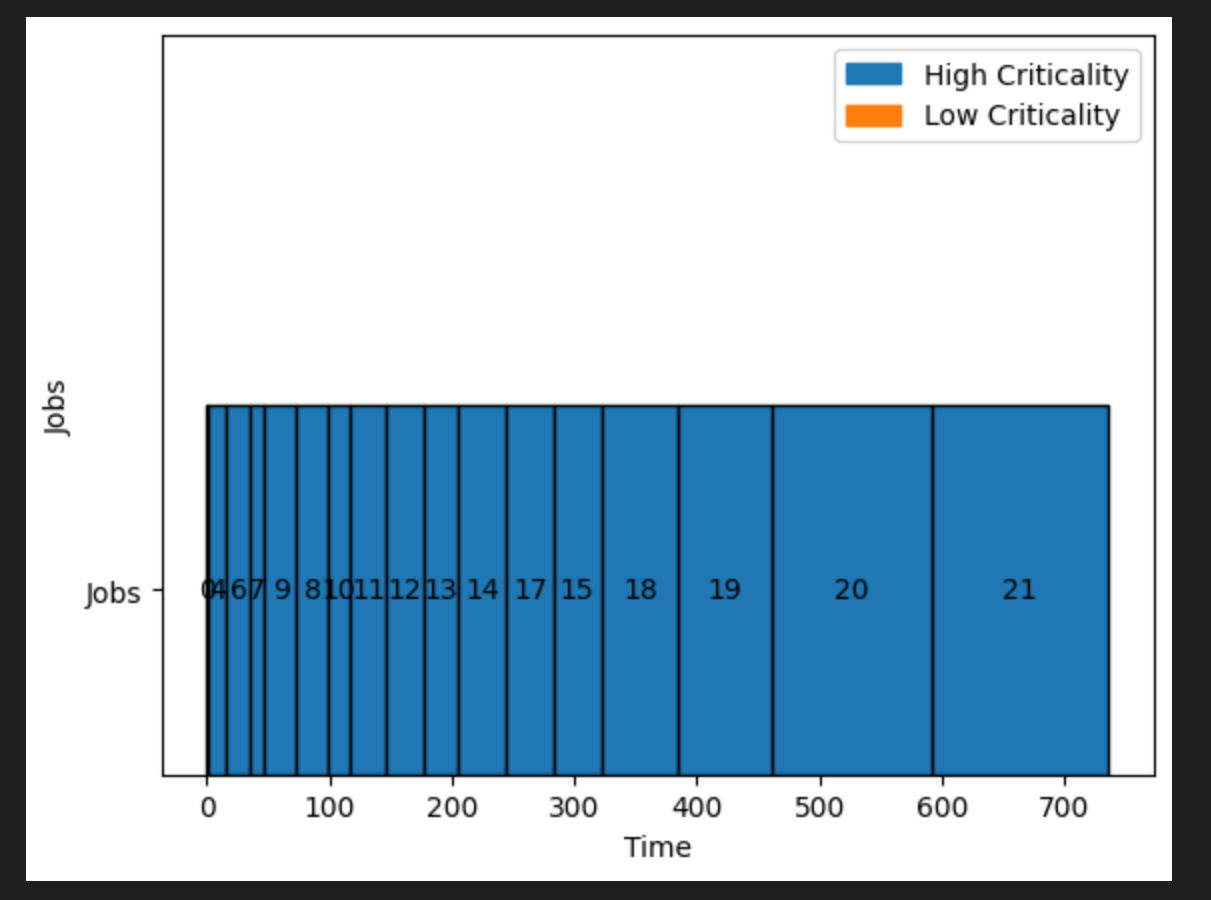}
  \caption{the schedule of an instance from Group 2}
  \label{fig:example}
\end{figure}
\paragraph{Group 1}
Doing the experiment on Group 1 instances, we could analyze the number of jobs it misses. Table 10 shows the results after doing these experiments. Results show that the processor can degrade to \textbf{51.02\%}

\begin{table}[!htbp]
\renewcommand{\arraystretch}{1.3}
\caption{The results of the experiment of Group 1 on varying speed processor}
\label{table_statistics_5}
\centering
\begin{tabular}{l|r|r}
\hline
\multicolumn{1}{c|}{} & \multicolumn{2}{c}{Statistics} \\
\hline\hline
 & Average job completion rate & Average missed jobs\\
\hline
HI jobs & 65\% & 1.6 jobs \\
Overall jobs & 68 \% & 2.608 jobs \\

\hline
\end{tabular}

\end{table}
\paragraph{Group 2}
Doing the experiment on Group 2 instances, we could analyze the number of jobs it misses. Table 11 shows the results after doing these experiments. Results also showed that the processor can degrade to \textbf{52.90 \%}

\begin{table}[!htbp]
\renewcommand{\arraystretch}{1.3}
\caption{The results of the experiment of Group 2 on varying speed processor}
\label{table_statistics_5}
\centering
\begin{tabular}{l|r|r}
\hline
\multicolumn{1}{c|}{} & \multicolumn{2}{c}{Statistics} \\
\hline\hline
 & Average job completion rate & Average missed jobs\\
\hline
HI jobs & 66\% & 3.05 jobs \\
Overall jobs & 65 \% & 7.7 jobs \\

\hline
\end{tabular}

\end{table}

\section{Sensitivity analysis}
\subsection{Varying the LO percentage}
We then do sensitivity analysis by varying the \textbf{LO percentage parameter} from 0.1 to 0.5 to see the effect of the percentage of Low criticality jobs on the behavior of the RL agent on both no degradation and varying speed processors. We test the agent on 1000 episodes for each value and report the average completion rate. 
\subsubsection{No degradation}
Figures 4 and 5 show the results of varying the LO percentage parameter under no degradation. The results show that our RL scheduler performed decently.
\begin{figure}[htbp]
  \centering
  \includegraphics[width=1\linewidth]{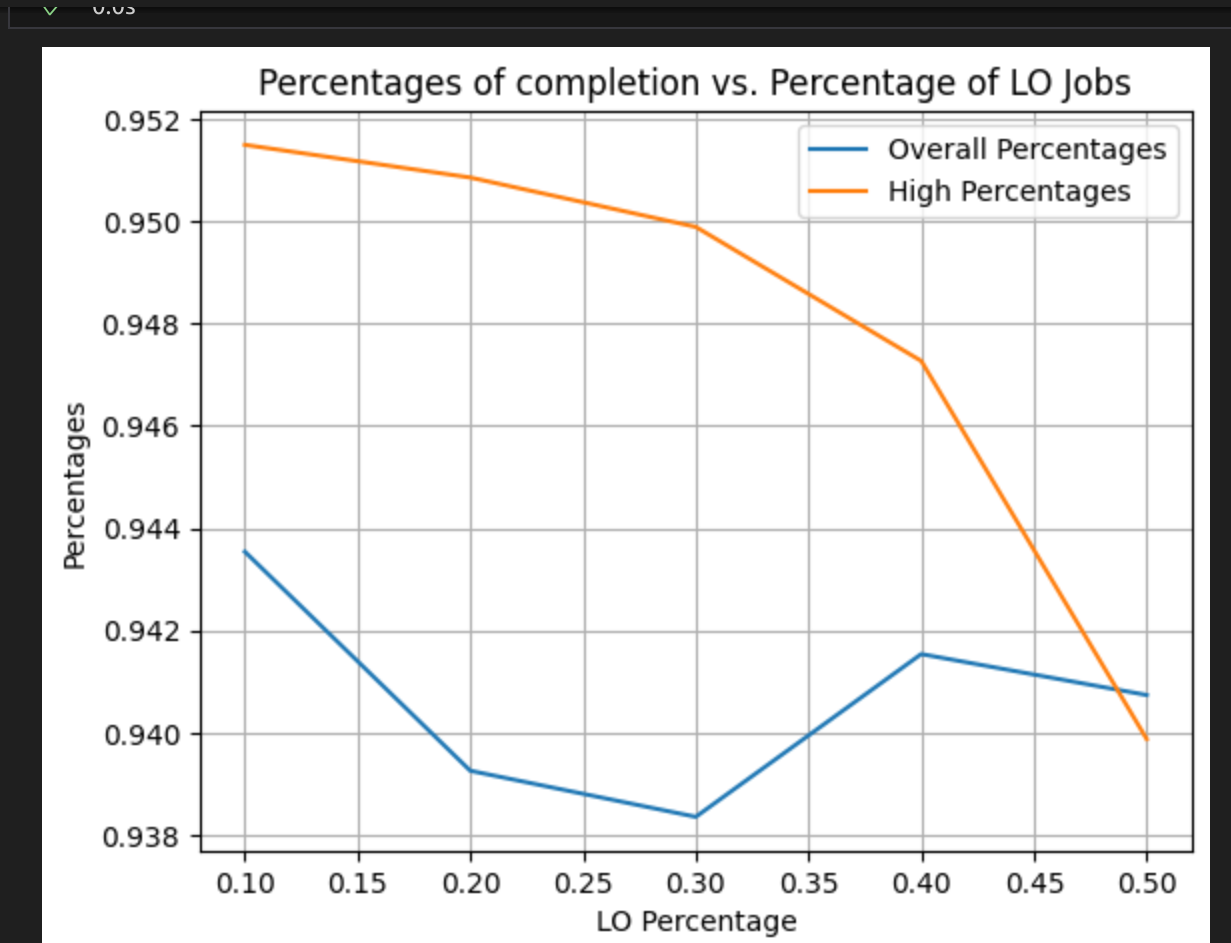}
  \caption{The average completion rate while varying the LO job percentage with no degradation.}
  \label{fig:example}
\end{figure}
\begin{figure}[htbp]
  \centering
  \includegraphics[width=1\linewidth]{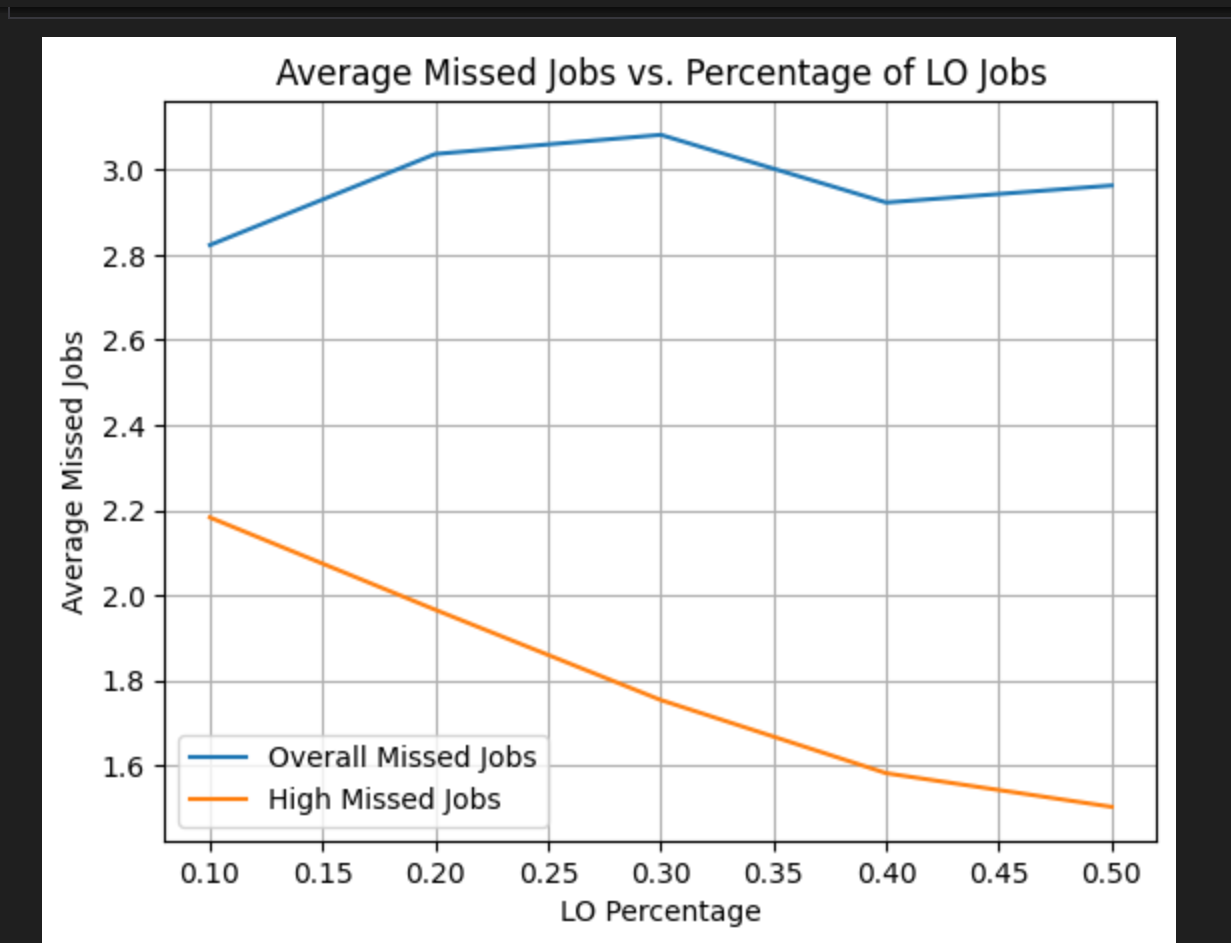}
  \caption{The average missed number of jobs while varying the LO job percentage with no degradation.}
  \label{fig:example}
\end{figure}

\subsubsection{With degradation}
We fixed the degradation threshold at \textbf{0.5} and varied the LO job percentage from 0.1 to 0.5 to see the effect of increasing the LO jobs. Figures 6 and 7 show the outcome of the agent scheduling under different LO job percentages. The results tend to show better performance with LO job percentage, i.e., higher HI jobs percentage, which agrees with the model goal to maximize the output of HI jobs under degradation.
\begin{figure}[htbp]
  \centering
  \includegraphics[width=1\linewidth]{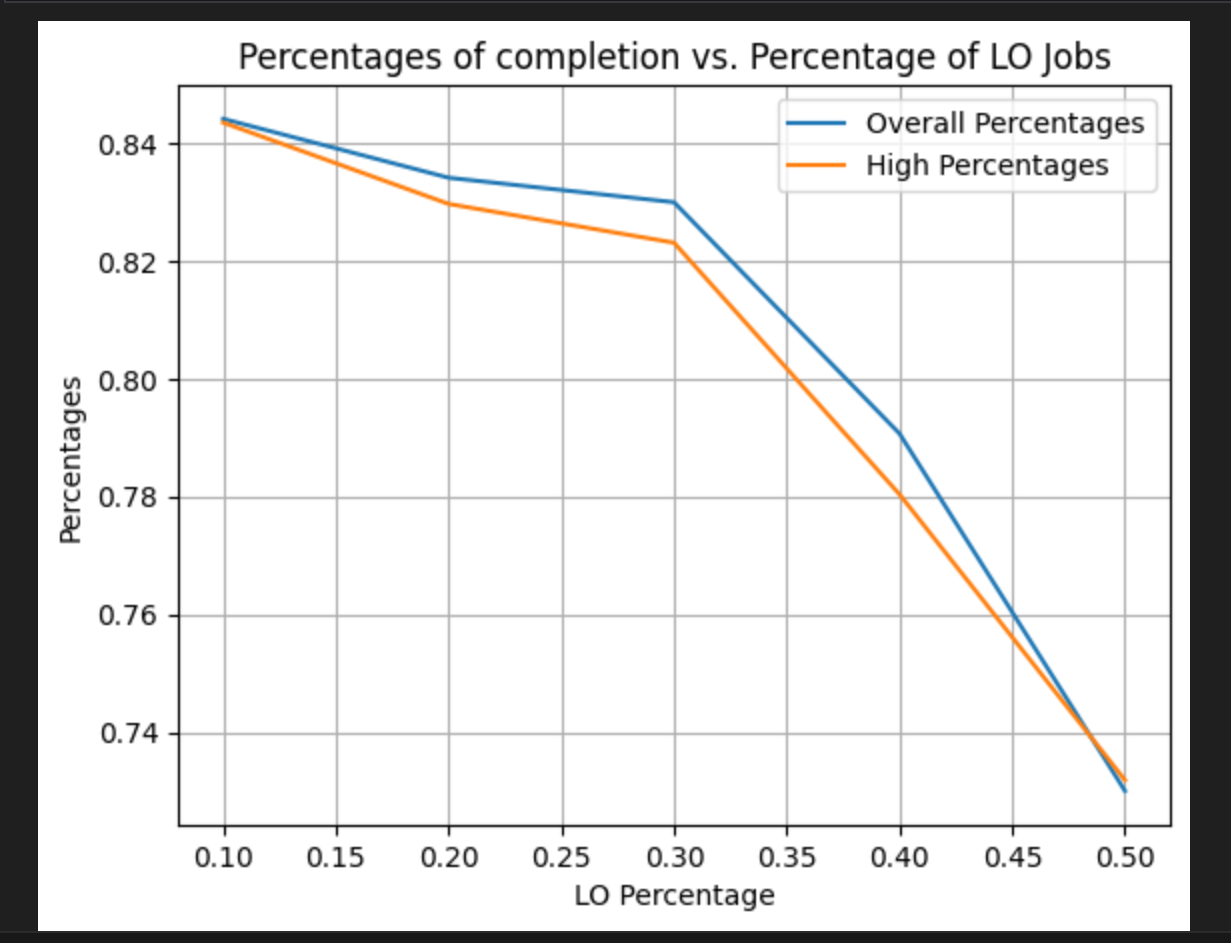}
  \caption{The average completion rate while varying the LO job percentage with degradation probability 0.5.}
  \label{fig:example}
\end{figure}
\begin{figure}[htbp]
  \centering
  \includegraphics[width=1\linewidth]{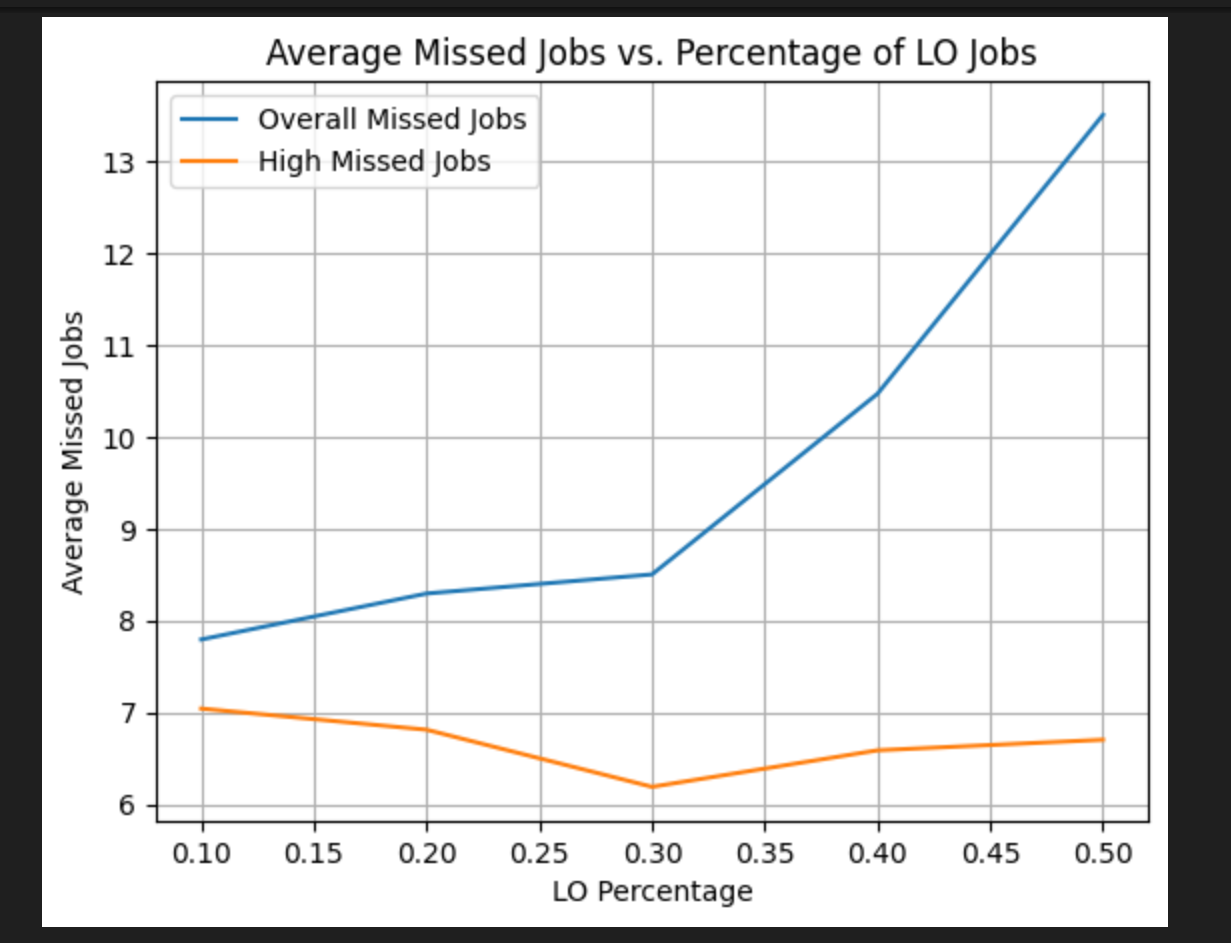}
  \caption{The average missed number of jobs while varying the LO job percentage with degradation probability 0.5.}
  \label{fig:example}
\end{figure}

\subsection{Varying the degradation threshold}
We varied the degradation probability threshold from 0.1 to 0.5 to see how the model behaves. Figure 8 shows the percentages with increasing degradation thresholds. While the agent doesn't seem to do that well overall with high threshold, it's part of how it was trained. As during degradation times, its focus should only be on HI criticality jobs. Additionally, Figure 9 shows the average number of missed jobs with increasing the degradation threshold.
\begin{figure}[htbp]
  \centering
  \includegraphics[width=1\linewidth]{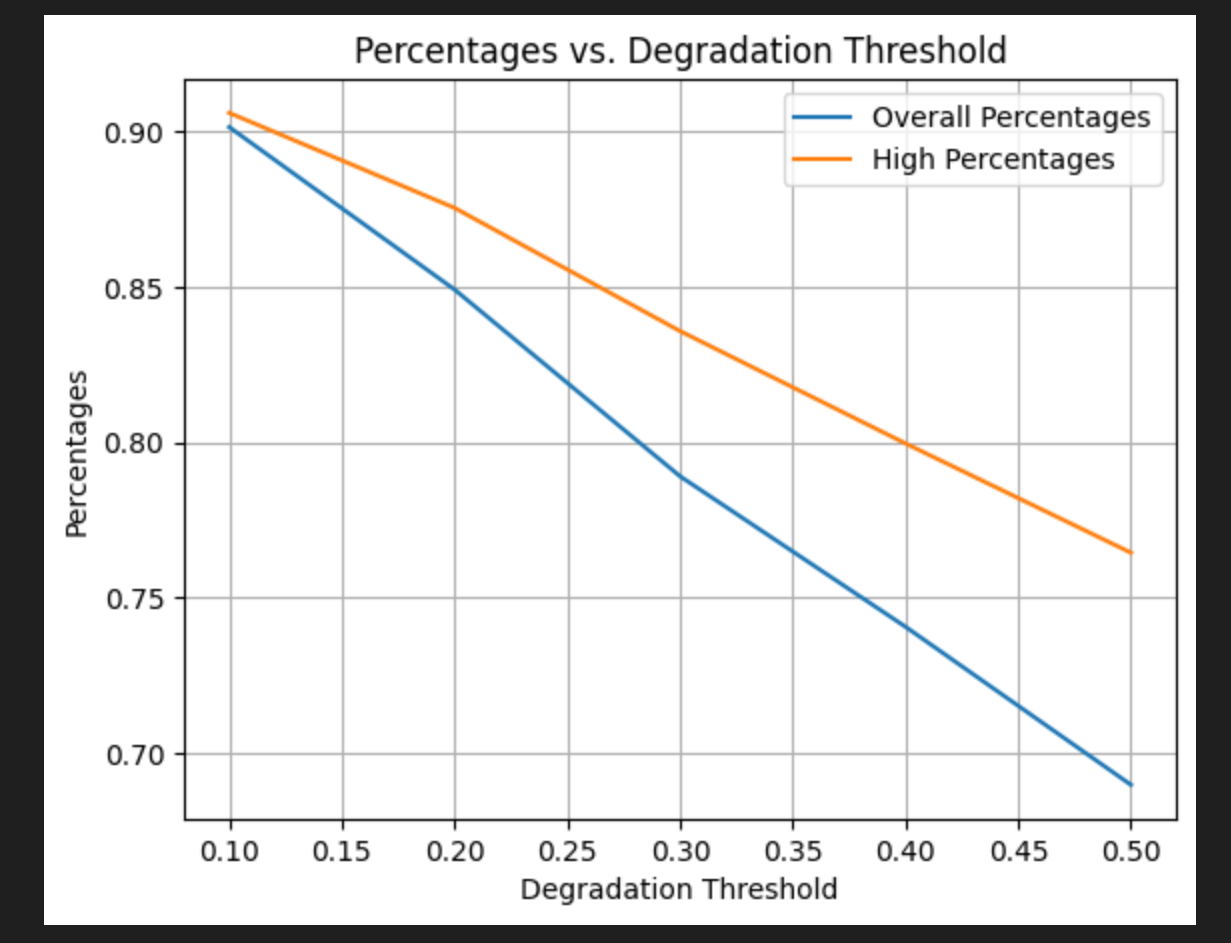}
  \caption{The average completion rate while varying the degradation probability.}
  \label{fig:example}
\end{figure}
\begin{figure}[htbp]
  \centering
  \includegraphics[width=1\linewidth]{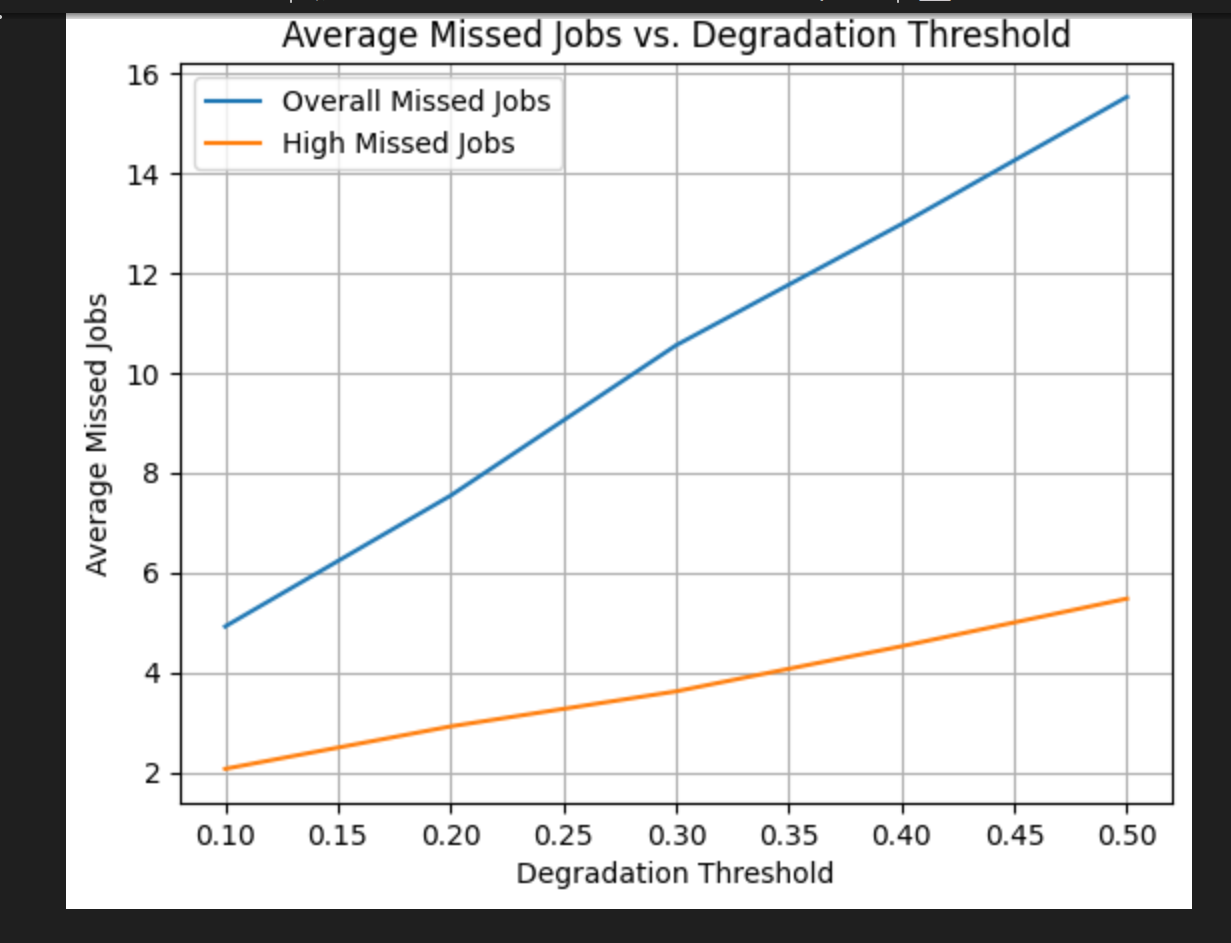}
  \caption{The average missed number of jobs while varying the degradation probability.}
  \label{fig:example}
\end{figure}

\section{Discussion}
The results of our work highlight the potential of reinforcement learning (RL) in addressing the complexities of mixed-criticality (MC) scheduling. The RL-based scheduler significantly improved the handling of diverse task sets with varying criticality levels, particularly in system degradation scenarios and varying processor speeds. This success can be attributed to the RL agent's ability to learn and adapt to the scheduling environment, optimizing for both high-critical task completion and overall system performance.

One key observation from our experiments is the RL agent's ability to prioritize tasks dynamically, adjusting to changes in task attributes such as release time, deadline, and processing time. This adaptability is crucial in real-time systems where conditions can change unpredictably. Moreover, using Markov Decision Processes (MDPs) provided a structured framework for modeling the scheduling problem, enabling the RL agent to make informed decisions based on the system's current state.

Additionally, our sensitivity analysis experiments show how the RL agent could prioritize the HI criticality jobs under the degradation conditions, maximizing their completion as much as possible.

However, a few challenges and limitations were identified. First, the computational overhead of training RL agents can be substantial, particularly for large-scale systems with numerous tasks. This necessitates further exploration into more efficient training algorithms and techniques to reduce computational costs. Additionally, while our approach performed well in synthetic and real server data scenarios, further validation in diverse real-world applications is necessary to confirm its generalizability and effectiveness.

Another area of interest is the impact of different RL algorithms on scheduling performance. While we employed a masked Proximal Policy Optimization (PPO) approach, exploring other algorithms, such as model-based RL, could provide deeper insights and potentially better performance.

Lastly, integrating safety constraints and fault tolerance into the RL framework remains an open question. Ensuring that the system can handle unexpected failures and maintain critical operations without compromising safety is paramount, especially in applications like autonomous vehicles or industrial automation.

Hence, while our RL-based scheduler for MC systems has shown promising results, several avenues for improvement and further research remain. Addressing these challenges will be crucial for RL's broader adoption and success in real-time and safety-critical scheduling applications.

\section{Conclusion and Future work}
In conclusion, the work tackles the dual criticality scheduling problem using reinforcement learning. This approach provided a more scalable and efficient solution. We demonstrated good results with synthetic data and real data from the server dataset studied in \cite{Carrasco2018:rcasEjor}, with low average missed jobs. We also do this work by assuming a processor with varying speeds that witnesses degradation. We finally suggest a good approach to extend this work and provide other variations to solve the MC scheduling problem below.

The results of this work have shown that RL is suitable for this type of problem. Providing scalable, efficient, and fast solutions for solving the NP-Hard problem of offline non-preemptive scheduling. Accordingly, this work can be extended to tackle the other variations of the MC scheduling problem, i.e., offline preemptive, online preemptive, and non-preemptive. In order to convert our work into a preemptive version, we suggest simple changes, including the details of the MDP design. For example, the time should be simulated so that the agent can choose a job every CPU cycle, even if it's not done with the current job. Additionally, the agent should be rewarded for choosing jobs of higher criticality and penalized for choosing lower ones. Furthermore, the agent's observation must include only the released jobs to convert this work into online scheduling. Finally, this work can be used to build more scalable schedulers of a higher number of jobs than what was proposed in this paper, 50 jobs. Additionally, it is possible to replace the uniform distribution model with a degradation forecasting module to mimic the CPU's real-time behavior.

\section*{Other Submissions}
This work was submitted to the 32nd International Conference on Real-Time Networks and Systems (RTNS 2024) on June 8, 2024.

\bibliographystyle{ACM-Reference-Format}
\bibliography{sample-base}

\end{document}